\documentclass[letterpaper]{article} 
\usepackage{aaai25}  
\usepackage{times}  
\usepackage{helvet}  
\usepackage{courier}  
\usepackage[hyphens]{url}  
\usepackage{graphicx} 
\usepackage{natbib}  
\usepackage{caption} 
\usepackage{amsmath}
\usepackage{algorithm}
\usepackage{booktabs} 
\usepackage{makecell}
\usepackage[noend]{algpseudocode}
\usepackage{newfloat}
\usepackage{listings}
\DeclareCaptionStyle{ruled}{labelfont=normalfont,labelsep=colon,strut=off} 
\floatstyle{ruled}
\newfloat{listing}{tb}{lst}{}
\floatname{listing}{Listing}
\title{TreeEval: Benchmark-Free Evaluation of Large Language Models through Tree Planning}

\author {
    Xiang Li\textsuperscript{\rm 1},
    Yunshi Lan\textsuperscript{\rm 1}\thanks{Yunshi is the corresponding author and she is also  affiliated with Shanghai Engineering Research Center of Big Data Management.},
    Chao Yang\textsuperscript{\rm 2}
}
\affiliations {
    \textsuperscript{\rm 1}East China Normal University\\
    \textsuperscript{\rm 2}Shanghai AI Laboratory\\
    xiang.li@stu.ecnu.edu.cn, yslan@dase.ecnu.edu.cn, yangchao@pjlab.org.cn
}

\usepackage{bibentry}

\begin{document}

\maketitle

\begin{abstract}
Recently, numerous new benchmarks have been established to evaluate the performance of large language models (LLMs) via either computing a holistic score or employing another LLM as a judge. 
However, these approaches suffer from data leakage due to the open access of the benchmark and inflexible evaluation process.
To address this issue, we introduce \textbf{TreeEval}, a benchmark-free evaluation method for LLMs that let a high-performance LLM host an irreproducible evaluation session and essentially avoids the data leakage. 
Moreover, this LLM performs as an examiner to raise up a series of questions under a topic with a tree planing strategy, which considers the current evaluation status to decide the next question generation and ensures the completeness and efficiency of the evaluation process.
We evaluate $6$ models of different parameter sizes, including $7$B, $13$B, and $33$B, and ultimately achieved the highest correlation coefficient with AlpacaEval2.0 using only around $45$ questions. 
We also conduct more analysis to show the robustness and reliability of TreeEval.
Our code can be accessed via the provided URL\footnote{https://github.com/Ashura5/TreeEval 

Due to AAAI's format limitations, the full version with appendices is available at \cite{li2024treeevalbenchmarkfreeevaluationlarge}.}.
\end{abstract}

\section{Introduction}
\label{sec:intro}

The recent surge in Large Language Models (LLMs) has been significant, transitioning from closed-source~\cite{openai2023gpt4,geminiteam2023gemini} to open-source~\cite{touvron2023llama,touvron2023llama2,jiang2023mistral} models. Various Supervised Fine-Tuning (SFT) and Reinforcement Learning from Human Feedback (RLHF) techniques have been proposed to further enhance the performance of LLMs~\cite{alpaca,vicuna2023,bai2022training,ouyang2022training,alignment_handbook2023}. These LLMs demonstrate capabilities to address diverse tasks and are widely utilized in both academic and industrial fields. While human evaluation is intuitive for assessing the performance of LLMs, it is time-consuming and susceptible to unexpected bias~\cite{zheng2023secrets,wang2024secrets}. Thus, investigating automatic evaluation approaches for LLMs becomes crucial.

To date, numerous automatic evaluation methods have been proposed. One approach involves annotating benchmark datasets, such as MMLU and BBH~\cite{hendrycks2021measuring,suzgun2022challenging}, to test various capabilities of an LLM. The performance is assessed by checking the overlap between annotated answers and generated answers, producing a holistic score to indicate the LLM's performance. We refer to this category of evaluation methods as the \textbf{benchmark paradigm}. However, the holistic score can be inflexible for measuring the quality of LLM outputs since token mismatches do not necessarily indicate incorrect answers.

With the advent of high-performance LLMs, another approach leverages them to simulate human evaluation. This involves providing the evaluated LLM with predefined benchmark questions and using another LLM, such as GPT-4, to judge its responses~\cite{zheng2023judging,alpaca_eval,bai2023benchmarking,wang2023large,zhang2023wider,wang2023pandalm,li2023generative,zhu2023judgelm}. We refer to this category of evaluation methods as the \textbf{LLM-as-judge paradigm}.
However, this evaluation approach can also introduce additional biases, including positional bias~\cite{wang2023large}, verbosity bias~\cite{saito2023verbosity}, and style bias~\cite{wu2023style}. Positional bias refers to the tendency to assign higher scores to answers based on their specific positions. Verbosity bias indicates that large language models often prefer more verbose answers, even if these longer responses are not necessarily of higher quality than shorter ones. Style bias manifests in the inclination of large language models to favor answers that match their own generated style, such as giving lower scores to correct responses with spelling errors, since LLMs rarely produce content with spelling mistakes.


\begin{figure}[t!]
    \centering
    \includegraphics[width=0.8\linewidth]{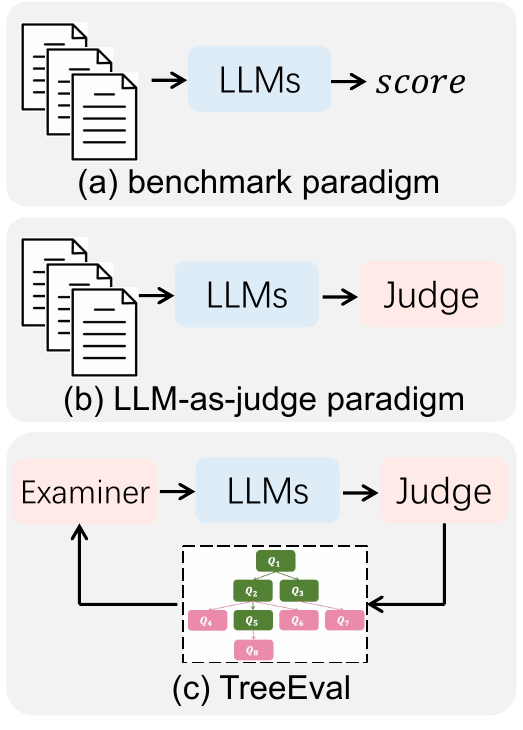}
    \caption{Comparison of TreeEval with existing evaluation paradigms. 
    }
    \label{fig:compare}
\end{figure}
Despite enabling automatic evaluation with standard pipelines, both the benchmark and LLM-as-judge paradigms face significant data leakage issues. The extensive training data used in LLM development, considered a valuable asset by many closed and even open-source models, can easily lead to benchmark data leakage, severely biasing evaluation results~\cite{zhou2023dont}.
To solve this issue, we propose a novel evaluation paradigm, which takes an LLM as an examiner to raise questions.
The examiner should produce different evaluation session for each time which makes it hard to duplicate the evaluation questions and protect the evaluation benchmark from disclosure for fine-tuning and pre-training an LLM deliberately.
However, simply adopting an LLM as examiner would lead to arbitrary evaluation question generation without a goal.
Designing such a benchmark-free evaluation method need take the following aspects into consideration:
(1) \textbf{Similar as the question in a benchmark}~\cite{alpaca,zheng2023judging}, the generated questions should be derived from certain topics, which ensures the scope of the evaluation.
(2) \textbf{Drawing inspiration from the interview}, within a topic, the examiner should generate a line of questions that are diverse to cover different knowledge rather than producing a single question.
(3) The generation procedure should be \textbf{flexible} enough to generate mutually connected questions and control the difficulty level of these questions.
When the current line of question cannot distinguish two LLMs, more difficult questions should be raised up.
Otherwise, the evaluation could be terminated immediately.

To this end, we propose \textbf{TreeEval}, which is a benchmark-free evaluation of the knowledge implication and question-answering capabilities of LLM through tree planning.
The line of questions within a topic for evaluation are organized in a tree, where each node contains a question.
In the process of constructing a tree, we repeatedly revisit the status of the current tree and generate the next node until the tree is enough to differentiate two LLMs.
The difference between our evaluation method and previous paradigms can be found in Figure~\ref{fig:compare}.
To verify the effect of our method, we evaluate multiple LLMs.
The results demonstrate that our method shows similar ranking as AlpacaEval2.0 in LLM-as-judge paradigm with only $45$ questions in average for each round of evaluation.
Further analysis shows our advantages in measuring fine-grained capabilities and conducting robust comparison for LLMs.

Our contributions are summarized as follows:
\begin{itemize}
\setlength\itemsep{-0.1em} 
\setlength\parskip{-0.1em} 
\item We introduce a novel evaluation paradigm, TreeEval, which allows for efficient and comprehensive evaluation of LLMs, inherently preventing data leakage issues.
\item TreeEval has advantage in distinguishing two LLMs with similar performance by constructing a deeper tree, which extends the evaluation process to obtain more stable and accurate assessment results.
\item We compare with a set of automatic evaluation baselines, and find that our TreeEval achieves the highest correlation coefficient with AlpacaEval2.0.
\end{itemize}
\section{Related Work}

\subsection{Methods of LLM Evaluation}

Due to the explosive growth and rapid update of LLMs, a significant challenge is to conduct accurate and comprehensive evaluation for them~\cite{chang2023survey}. 
Early studies leverage open-ended question answering datasets and math word problems as the evaluation benchmarks~\cite{touvron2023llama,anil2023palm,chen2024chatgpts}
to evaluate the commonsense knowledge and reasoning capabilities of LLMs.
Subsequently, more benchmark datasets like MMLU~\cite{hendrycks2021measuring}, AGIEval~\cite{zhong2023agieval}, IFEval~\cite{zhou2023instructionfollowing}
have been elaborately designed to gauge diverse abilities of LLMs.
Some studies~\cite{wang2023large,wang2023large,saha2023branchsolvemerge} go beyond standard evaluation metrics.
They evaluate the quality and accuracy of predicted results through human annotation, which is able to provide a more comprehensive feedback.
With the emergence of high-performance LLMs like GPT-4~\cite{openai2023gpt4}, Gemini Pro~\cite{geminiteam2023gemini}, more recent studies start to utilize them to simulate the human evaluation process.
In this realm, PandaLM~\cite{wang2023pandalm} strives to provide reproducible and automated comparison between various LLMs by training a LLM as the judge.
GPTScore~\cite{fu2023gptscore} and G-Eval~\cite{liu2023geval} utilize GPT-3 and GPT-4 as the judge to evaluate the LLMs with incorporation of in-context learning and chain-of-thought strategies.
The above methods rely heavily on a well-organized benchmark dataset. However, there have been some recent works focusing on data leakage of LLM reviews. \cite{zhu2024dyval} proposed a method based on DAG to dynamically generate samples to evaluate LLM reasoning capabilities during the evaluation process.
And our method is benchmark-free and has LLMs performing as the examiner to evaluate other models’ knowledge entailment and question answering capabilities.

\subsection{Data Leakage of LLM Evaluation}

As the number of benchmarks for language model evaluation increases, data leakage emerges as an inevitable concern. However, there appear to be a limited number of studies addressing this issue.
\citeauthor{sainz-etal-2023-nlp}(\citeyear{sainz-etal-2023-nlp}) propose a method to detect data breaches in closed-source LLMs, based on the premise that LLMs can recall training data and tend to reproduce similar content. 
\citeauthor{zhou2023dont}(\citeyear{zhou2023dont}) conduct qualitative analysis of the impact of data leakage, which suggests that a data breach in one benchmark significantly enhances the LLM's performance on that specific benchmark while diminishing its capabilities on other uncompromised benchmarks. 
\citeauthor{yang2023rethinking}(\citeyear{yang2023rethinking}) propose a more accurate approach which employs an LLM detector with top-k closest training data to determine if they match the test data.
In contrast to these methods, which develop additional models for detecting data leakage during LLM evaluation with given benchmark datasets, our proposed method introduces a novel paradigm for LLM evaluation. It not only ensures the high quality of test questions but also inherently avoids data leakage.

\section{Methodology}
\label{sec:method}

\begin{figure*}[t]
    \centering
    \includegraphics[width=\linewidth]{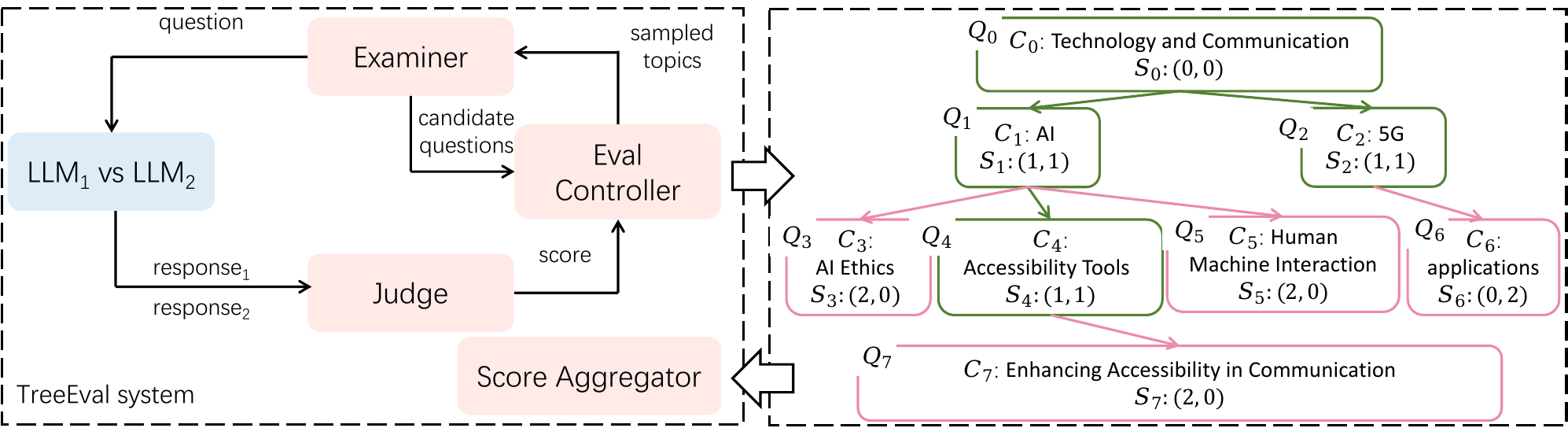}
    \caption{TreeEval system with an illustrative tree for evaluation.
    The left section contains the components and their workflow in TreeEval. The right section displays a constructed tree within topic \textit{Technology and Communication} for evaluation (the leaf nodes are shown in red boxes), where each node denotes a question annotated with its topic and evaluation score.
    We further display the generated questions of the tree in the (Appendix Eval Controller Example).}
    \label{fig:method}
\end{figure*}

\subsection{Overall Architecture}

Figure~\ref{fig:method} shows the overall structure of TreeEval. TreeEval organizes evaluations in a tree format, using components such as an \textit{Examiner}, a \textit{Judge}, and an \textit{Eval Controller}. After the tree is built, an \textit{Aggregator} compiles the scores. This framework allows for benchmark-free evaluation of LLMs through tree planning.
Here's how it works:
\begin{enumerate}
\setlength\itemsep{-0.1em} 
\setlength\parskip{-0.1em} 
    \item \textbf{Session Setup}: For each evaluation session, we choose two LLMs and start with an initial topic.
    \item \textbf{Question Generation}: The Examiner generates questions within this topic.
    \item \textbf{Response Collection}: These questions are sent to the LLMs, and their responses are collected.
    \item \textbf{Response Evaluation}: The Judge compares the responses and decides the winner for each question.
    \item \textbf{Evaluation Control}: If the responses are closely matched, the Eval Controller deepens the question. If a clear winner is found, the process moves to a new question. This follows a breadth-first search strategy, ensuring diverse and reliable questions.
    \item \textbf{Score Aggregation}: Finally, the Aggregator compiles the scores from all nodes in the tree to produce a comprehensive evaluation score.
\end{enumerate}
TreeEval uses a tree structure to evaluate LLMs, minimizing the number of questions needed. Questions are generated automatically, preventing benchmark leakage. The root node starts with the session's topic, and each node represents a question within that topic. Connections between nodes show how questions evolve. Deeper nodes indicate more similar abilities between the LLMs. Sibling nodes, derived from the same parent, cover different subtopics of the same main topic.

\subsection{TreeEval Modules}

In this section, we provide more details of the components of the TreeEval and illustrate how to construct a tree for evaluation via these components.

\noindent \textbf{Examiner.}
The examiner is a LLM-based module, which takes charge of generating exam questions that are able to cover diverse topics.
Following \cite{bai2023benchmarking}, we pre-define a set of topics 
as the scope of evaluation.

As the initialization of an evaluation session, we randomly sample a topic from the pre-defined topic set, which is denoted as $\mathcal{FC}_{\text{pre-define}}$.
Given a topic, the examiner is requested to craft a question that related to it via a prompt with the consideration of the coherence to the topic and the required format of the question.
The detailed instruction is displayed in (Appendix Prompt for Examiner).


Once the session begins, we organize the follow-up questions in a tree structure.
For simplify, we generally denote the follow-up topic at the $t$-th time step as $C_t$.
And the above procedure can be presented as:
\begin{align*}
Q_t = \text{Examiner}(C_t).
\end{align*}
Subsequently, $Q_t$ is utilized as the question to test the LLMs under review.



\noindent \textbf{Judge.}
Previous studies~\cite{wang2023large} conduct pairwise comparison and identify the superior responses among two evaluated LLMs, which has advantage in providing more nuanced assessment.
Following these studies, we consult a pair of LLMs with the same question.
The detailed instruction is displayed in (Appendix Prompt for Judge).
After the responses have been produced via the LLMs, another LLM performs as the judge to the responses.


To ensure the reliability of the judge, we further conduct exchange evaluation, that is to switch the order of the responses.
This procedure can be denoted as:
\begin{align*}
S_t^{1} & = \text{Judge}(Q_t, A_t^1, A_t^2); \\
S_t^{2} & = \text{Judge}(Q_t, A_t^2, A_t^1), 
\end{align*}
where $A_t^1$, $A_t^2$ denote the responses from the pair of LLMs for $Q_t$.
Each output judges the winner is $A_t^1$ or $A_t^2$ or a tie exits.
If there is an agreement for $S_t^1$ and $S_t^2$, We assign $2$ score to the winner and $0$ score to the loser to form $S_t$.
Otherwise, we assign $1$ to each model as $S_t$. 



As the evaluation proceeds, we maintain a memory to record the history of the session, including the initial topic, historical questions as well as responses from the two evaluated LLMs.
After $Q_t$ has been responded, the history at the $t$-th time step in the evaluation session can be denoted as $\mathcal{M}_{t} = \{C_0, Q_0, A_0^1, A_0^2, ..., C_t, Q_{t}, A_{t}^1, A_{t}^2 \}$.
To involve the coherence of the flowing conversation and raise up rational follow-up questions, we prompt the examiner with the consideration of the history.

\begin{algorithm*}
\caption{Procedure of TreeEval}
\begin{algorithmic}[1] 
\small
\State \textbf{Input} $\mathcal{FC}_{\text{pre-define}}$;
\State \textbf{Initial} $t \leftarrow 0$; $\mathcal{M}_t \leftarrow \emptyset $ 
\While{Termination strategy is not satisfied}
    \For {$C_t \in \mathcal{FC}_{parent}$ or $C_0 \in \mathcal{FC}_{\text{pre-define}}$}
        \State $\Tilde{\mathcal{Q}}_t \leftarrow \text{Examiner}(C_t)$ \Comment{Sample questions via Examiner.}
        \State $Q_t \leftarrow \arg \max_{Q^i_t \in \Tilde{\mathcal{Q}}_t}(\text{Sim}(Q^i_t, C_t) - \max_{Q_k \in \mathcal{M}_t} \text{Sim}(Q^{i}_{t}, Q_k))$ \Comment{Rank candidate questions in Step Two.}
        \State $A^1_t, A^2_t \leftarrow \text{LLMs} (Q_t)$
        \State $S_t = \text{Judge}(Q_t, A^1_t, A^2_t)$ \Comment{Output scores of LLMs via Judge.}
        \State $\mathcal{M}_t \leftarrow \mathcal{M}_t \cup \{C_t, Q_t, A^1_t, A^2_t\}$ 
        \State $\Tilde{\mathcal{FC}}_{t} \leftarrow \text{NER}(A_t^1) \cup \text{NER}(A_t^2)$ \Comment{Generate candidate topic in Step Two.}
        \State $\mathcal{FC}_{t} \leftarrow \emptyset$
        \While{$|\mathcal{FC}_{t}| < k$} \Comment{Iteratively Filter candidate topics in Step One.}
            \State $C^i_{t} \leftarrow \arg \max_{\Tilde{C}^i_{t} \in \Tilde{\mathcal{FC}}_{t}} (\text{Sim}(\Tilde{C}^i_{t}, C_t))$
            \State $\mathcal{FC}_{t} \leftarrow \mathcal{FC}_{t} \cup \{C^i_{t} \}$
            \State $\Tilde{\mathcal{FC}}_{t} \leftarrow \Tilde{\mathcal{FC}}_{t} \setminus C^i_{t}$
            \For {$\Tilde{C}^j_{t} \in \Tilde{\mathcal{FC}}_{t}$}
                \State $\text{Sim}(\Tilde{C}^j_{t}, C_t) \leftarrow \text{Sim}(\Tilde{C}^j_{t}, C_t) - \text{Sim}(\Tilde{C}^j_{t}, C^i_{t})$
            \EndFor
        \EndWhile
        \State $t \leftarrow t + 1$
    \EndFor
\EndWhile
\end{algorithmic}
\label{ag:control}
\end{algorithm*}

\noindent \textbf{Eval Controller.}
The evaluation controller takes charge of the process of tree planning.
Arbitrary generation of questions result in unorganized evaluation of LLMs with repeated questions and limited topics.
To ensure the relevance and diversity of the generated questions, we have the following consideration: 
(1) To simulate the real-world interview of a certain subject, where the questions in an examination are mutually connected, we assume the generated follow-up question should be closely linked to its previous question via topics.
For example, in Figure~\ref{fig:method}, inheriting from the root topic ``\textit{technology and communication}'', we can raise a question on ``\textit{5G}'' that is relevant to the root topic and goes deeper.
(2) The generated questions should not be repeated in the existing questions and we should ensure the diverse knowledge covered by the tree.
For example, in Figure~\ref{fig:method}, under the topic ``\textit{AI}'', we can come up with distinct but related sub-topics as siblings such as ``\textit{AI Ethics}'', ``\textit{Accessibility Tools}'' and ``\textit{Human Machine Interaction}''.


Inspired by the Tree-of-Thought~\cite{long:arxiv2023}, where a controller produces the next thought step, we let Eval Controller arrange the follow-up evaluation according to $\mathcal{M}_{t}$.
On the one hand, it prepares the follow-up topics $\mathcal{FC}_{t}$ based on $\{C_t, A^1_t, A^2_t\} \in \mathcal{M}_{t}$ for any of its child nodes in advance.
On the other hand, it determines $Q_{t+1}$ based on the $\mathcal{FC}_{t}$ and $\{Q_1, Q_2, ..., Q_t\} \in \mathcal{M}_{t}$ if the $t$-th node is the parent node at $t+1$ time step.
We next describe the above two steps in detail:
\begin{itemize}
\setlength\itemsep{-0.1em} 
\setlength\parskip{-0.1em} 
    \item \textbf{Step One}: Sample topics from the responses of the previous question: $\mathcal{FC}_{t} \sim \text{NER}(A_t)$\footnote{The detailed instruction could be found in (Appendix Prompt for NER)}.
    This works better when the Named Entity Recognition (NER) tool is built upon a LLM as some relevant entities could be revised via the model instead of solely being extracted~\cite{wang:arxiv2023}. 

    We sample candidate topics from both $A_t^1$ and $A_t^2$ then merge them together, which results in a set of candidate topics $\tilde{\mathcal{FC}}_{t}$ as the follow-up topics of $t$-th node.
    However, this may produce some candidates that are repeated.
    To avoid this, we first measure the similarity between $C^{i}_{t} \in \tilde{\mathcal{FC}}_{t}$ and $C_t$ by computing the Cosine Similarity of their encoded vector representation~\cite{zhang2023retrieve}, which is denoted as $\text{Sim}(C^{i}_{t}, C_t)$.
    Then, we iteratively push out $C^{i}_{t}$ with the largest score.
    Next, we update the similarity scores of the rest topic $C^{j}_{t}$ by subtracting the similarity score of $C^{j}_{t} \in \tilde{\mathcal{FC}}_{t} \setminus C^{i}_{t}$ and $C^{i}_{t}$, which is to decrease the possibility of retrieving similar topics.
    This procedure continues until we have pushed out $k$ topics as $\mathcal{FC}_{t}$ for the follow-up question generation.

    \item \textbf{Step Two}: If the question at $(t+1)$-th time step is the child node of the node at $t$-th time step, we generate questions based on the sampled topic via $Q^{i}_{t+1} \sim \text{Examiner}(C_{t+1})$, where $C_{t+1} \in \mathcal{FC}_{t}$.
    This could form a candidate question set $\tilde{\mathcal{Q}}_{t+1}$.
    Still, to avoid repetition of the generated questions and ensure a broad spectrum of inquiry questions, we conduct ranking for the candidate questions.
    Specifically, we measure the similarity between $Q^{i}_{t+1} \in \tilde{\mathcal{Q}}_{t+1}$ and $C_{t+1}$ via Cosine Similarity.
    Then we push out $Q^{i}_{t+1}$ with the largest similarity score of $\text{Sim}(Q^{i}_{t+1}, C_{t+1})$ and the least similarity score of $\arg \min_{Q_k \in \mathcal{M}_t} \text{Sim}(Q^{i}_{t+1}, Q_k)$.
\end{itemize}



\noindent \textbf{Termination Strategy.}
To determine whether we should stop generating subsequent questions along a topic, we identify several termination criteria:

\begin{itemize}
\setlength\itemsep{-0.1em} 
\setlength\parskip{-0.1em} 
    \item For each node in the tree, if the question posed by the current topic successfully distinguish the capability of the two LLMs under review. 
    Alternatively, if there is no tie for the current question, we terminate the child node search under the current node.
    \item After we have generated the sibling nodes of a parent node, we revisit the scores of these siblings. 
    If it shows a dominate score over all these siblings, this indicates that we have a winner for this branch. 
    Hence we stop further search for any of these sibling nodes.
    \item To prevent the evaluation session preceding indefinitely, a maximum depth $T$ for the tree search is pre-defined.
    Once the limit reaches, we terminate the child node search under the current topic.
\end{itemize}





We terminate a tree search when every node in the tree satisfies the above criteria.
The entire process is described in Algorithm~\ref{ag:control}.

\subsection{Score Aggregator}


After we have constructed the multiple trees across $\mathcal{FC}_{\text{pre-define}}$, where the nodes in each tree implies the win-rate between two LLMs under review towards a specific topic.
To yield a final win-rate result, we aggregate the scores of these constructed trees.
However, it is irrational to consider all the nodes in a tree equally due to their different features and result scores.
Specifically, we take the following aspects of $t$-th node in a tree into account when we aggregate their scores:

\begin{itemize}
\setlength\itemsep{-0.1em} 
\setlength\parskip{-0.1em} 
    \item \textbf{Distance to the root node.} 
    Based on the principle of an evaluation session, a longer distance to the root node indicates a more intensive competition between the evaluated LLMs and the more important the node is.
    This suggests that the winner only has a marginal advantage over the other one. 
    Therefore, we define one aspect of an important node as $w^{\text{root}}_t = \frac{1}{d}$, where $d$ is the distance from the $t$-th node to the root node in a tree.
    \item \textbf{Origin of the topic.} 
    As the topic is derived from the responses in its parent node, a node inherited the topic generated from responses of the losing LLM is more important considering it is more likely to  balance the situation. 
    Hence, we define one aspect of an important node as:
    \begin{align*}
       w^{\text{topic}}_t = \begin{cases} 
       1 & \text{Topic originated from the loser} \\
       0.5 & \text{Otherwise}
       \end{cases}
    \end{align*}
    \item \textbf{Variance of the sibling nodes.}
    The disagreement of the evaluation of the sibling node may implicit a potential randomness derived from the topic. 
    So we define the sibling consensus as:
    \begin{align*}
       w^{\text{topic}}_t = \frac{1}{\sigma^2 + 1},
    \end{align*}
    where $\sigma$ is the variance of the score of its sibling nodes. 
\end{itemize}

Considering the above aspects, we compute the final importance weights of $t$-th node as:
\begin{align*}
   w_t = {w^{\text{root}}_t}^{\alpha} \cdot {w^{\text{topic}}_t}^{\beta} \cdot {w^{\text{sibling}}_t}^{\gamma},
\end{align*}
where $\alpha$, $\beta$, and $\gamma$ are hyper-parameters indicating the relative importance of these aspects.
As a result, we sum up the $w_t$ multiplying with the win-rate of an LLM and devide the total evaluation questions to obtain its final scores:
\begin{align*}
   S = \frac{1}{N}&\sum_{i\text{-th Tree from } \mathcal{FC}_{\text{pre-define}};} \sum_{t\text{-th node in } i\text{-th Tree}} w_t \cdot S_t,
\end{align*}
where $N$ is the sum of node weights in the evaluation session and $S$ is normalized. 
\section{Experiments}
\label{sec:exp}

\begin{table*}[t!]
\captionsetup{aboveskip=1pt, belowskip=1pt} 
\centering
\small
\begin{tabular}{l | c | c | c | c | c | c c }
\toprule
LLMs & \thead{MMLU}$^\star$ & \thead{BBH}$^\star$ & \thead{AlpacaEval}$^\dagger$ & \thead{MT-bench}$^\dagger$ & \thead{AlpacaEval2.0}$^\dagger$ & \multicolumn{2}{c}{TreeEval(Ours)} \\
\midrule
  & Acc & Acc & Win-Rate & score & Win-Rate& \#Q & Score(var)  \\
\midrule
\texttt{Mistral-7B-Instruct-v0.2} & $70.6$ & $46.4$ & $92.78$ & $8.30$ &$14.72$  & $-$ & $2.50(0.000)$  \\ %
\midrule
\texttt{Yi-34B-Chat} & $73.46$ & $71.74$ & $94.08$ & $8.65$ & $27.19$ & $31.67$ & $3.48(0.011)$  \\ %
\midrule
\texttt{xwinlm-13b-v0.1} & $56.6$ & $37.58$ & $91.76$ &  $7.34$ & $17.43$ & $62.33$ & $2.67(0.000)$  \\ %
\midrule
\texttt{WizardLM-13B-V1.2} & $52.7$ & $40.12$ & $89.17$ & $7.2$ & $12.03$& $44.67$ & $1.10(0.070)$ \\ %
\midrule
\texttt{zephyr-7b-beta} & $61.4$ & $42.72$ & $90.60$ & $7.34$ &$10.99$ & $45.67$ & $2.19(0.003)$  \\ %
\midrule
\texttt{Vicuna-33b-v1.3} & $59.2$ & $52.0$ & $88.99$ & $7.12$ &$12.71$ & $41.33$ & $1.61(0.044)$  \\ %
\midrule

\midrule
Average \#Q $\downarrow$ & $14,079$ & $6,511$ & $804$ & $80$ & $804$ & $\mathbf{45.1}$ & $-$ \\
\midrule
$\rho$ $\uparrow$ & $0.43$ & $0.37$ & $0.71$ & $0.61$ & $1.0$ & $-$ & $\mathbf{0.83}$\\
\midrule
$\tau$ $\uparrow$ & $0.33$ & $0.33$ & $0.47$ & $0.41$ & $1.0$ & $-$ & $\mathbf{0.73}$\\
\bottomrule
\end{tabular}
\caption{Comparison of LLMs across various evaluation methods. 
``$\star$'' denotes we re-implement MMLU and BBH benchmarks~\cite{chia2023instructeval}, calculating results in both $5$-shot and $3$-shot contexts. 
``$\dagger$'' denotes we directly take results from the respective leader-boards from MT-bench, AlpacaEval, and AlpacaEval2.0. 
``\#Q'' denotes the number of questions used for evaluation.
We report the correlation of rankings obtained through different methods with those from AlpacaEval2.0, using $\tau$ for the Kendall correlation coefficient~\cite{10.1093/biomet/30.1-2.81} and $\rho$ for the Spearman correlation coefficient~\cite{ca468a70-0be4-389a-b0b9-5dd1ff52b33f}.} 
\label{tab:llm_comparison}
\end{table*}

\subsection{Experimental Setup}
\noindent \textbf{Evaluated LLMs.} We evaluated the following open-source LLMs, including two 7B models, two 13B models, and two 33B models. These models are either derived from LLaMA~\cite{touvron2023llama,touvron2023llama2} or trained from scratch using the LLaMA architecture, and some show similar performance according to the open-source LLM leaderboard\footnote{\url{https://tatsu-lab.github.io/alpaca_eval/}}.

\begin{itemize}
\setlength\itemsep{-0.1em} 
\setlength\parskip{-0.1em} 
    \item \texttt{Yi-34B-Chat}~\cite{Yi34BChat2023} is a product from 01.AI, built on a large-scale multilingual dataset.
    \item \texttt{Xwin-LM-13B-V0.1}~\cite{xwin-lm} is based on LLaMA2-13B and tuned through SFT and RLHF.
    \item \texttt{Mistral-7B-Instruct-v0.2}~\cite{jiang2023mistral} is tuned on the Mistral-7B model, built with the LLaMA architecture.
    \item \texttt{Vicuna-33B-v1.3}~\cite{zheng2023judging} originates from LLaMA-33B and is fine-tuned using dialogues from ShareGPT.
    \item \texttt{WizardLM-13B-V1.2}~\cite{xu2023wizardlm} is based on LLaMA2-13B and fine-tuned with enhanced instruction data using Evol-Instruct.
    \item \texttt{Zephyr-7B-beta}~\cite{tunstall2023zephyr} is derived from Mistral-7B and aligned using SFT and DPO methods.
\end{itemize}
\noindent \textbf{Comparable Evaluation Methods.} We compare TreeEval with several existing methods, including:

\begin{itemize}
\setlength\itemsep{-0.1em} 
\setlength\parskip{-0.1em} 
    \item \textbf{Benchmark Paradigm}:
    \begin{itemize}
    \setlength\itemsep{-0.1em} 
    \setlength\parskip{-0.1em} 
        \item \textbf{MMLU}~\cite{hendrycks2021measuring}
        \item \textbf{Big-Bench Hard (BBH)}~\cite{suzgun2022challenging}
    \end{itemize}
    \item \textbf{LLMs as Judges}:
    \begin{itemize}
    \setlength\itemsep{-0.1em} 
    \setlength\parskip{-0.1em} 
        \item \textbf{AlpacaEval} and \textbf{AlpacaEval2.0}~\cite{alpaca_eval}
        \item \textbf{MT-Bench}~\cite{zheng2023judging}
    \end{itemize}
\end{itemize}

AlpacaEval and AlpacaEval2.0 use ChatGPT as the judge for single-turn interactions, while MT-Bench focuses on multi-turn dialogues.

\noindent \textbf{Implementation Details.} We use \texttt{GPT-4-0613} as the examiner, deployed with FastChat~\cite{zheng2023judging}. The temperature is set to 1 to generate varied questions. We set $T$ and $k$ to 3. Parameters $\alpha$, $\beta$, and $\gamma$ are set to 1, 1, and 0.4, respectively. To reduce randomness, we repeat the experiments three times, average the scores, and additionally report the variance across runs to demonstrate that our method enables more stable comparisons between models with similar performance. We use \texttt{Mistral-7B-Instruct-v0.2} as the reference model for pairwise comparison due to its moderate performance on public leaderboards.

\subsection{Performance of TreeEval}

We display the performance of TreeEval in Table~\ref{tab:llm_comparison}, from which we have the following observations:
(1) Among all the comparable evaluation methods, our method is able to achieve the highest correlation coefficient with the rankings of AlpacaEval2.0 on the indicators of both $\rho$ and $\tau$.
AlpacaEval2.0 is commonly viewed as the recognized LLM evaluation leader-board and the high consistency between our ranks indicates the reliability of our method.
(2) Our method is able to complete the evaluation procedure with only $45$ questions in average while the other evaluation methods require much more questions to generate an evaluation result.
This indicates that our evaluation is efficient on evaluating LLMs with minimum questions.
(3) Since we treat \texttt{Mistral-7B-Instruct-v0.2} as the reference for the pairwise comparison, we notice the larger gap between the evaluated LLM and the reference is, the less test questions are proposed in the evaluation session, which shows that tree planning indeed meets our expected motivation.
We further display the pairwise correlation in (Appendix More analysis) to show the correlation between TreeEval and AlpacaEval is also high.

\subsection{Further Analysis}
\label{more analysis}

We further analyze to verify TreeEval's effect.\footnote{
Due to the page limit, more analyses (i.e., \textbf{Fine-grained Evaluation} and \textbf{robustness of TreeEval}) are provided in (Appendix Fine-grained Evaluation) and (Appendix Robustness of TreeEval), respectively.}

\begin{table*}[t!]
\captionsetup{aboveskip=2pt, belowskip=2pt} 
\centering
\small
\begin{tabular}{l | c | c | c | c | c c }
\toprule
LLMs & \thead{MMLU}$^\star$ & \thead{BBH}$^\star$ & \thead{MT-bench}$^\dagger$ & \thead{AlpacaEval2.0}$^\dagger$ & \multicolumn{2}{c}{TreeEval(Ours)} \\
\midrule
  & Acc & Acc & score & Win-Rate& \#Q & Score(Var)  \\
\midrule
\texttt{Yi-34B-Chat} & $73.5$ & $71.7$ & $8.65$ &$27.1$  & $-$ & $2.50(0.000)$  \\ %
\midrule
\texttt{Qwen1.5-110B-Chat} & $80.4$ & $74.8$ & $8.88$ &$33.7$  & $42.67$ & $4.03(0.110)$  \\ %
\midrule
\texttt{Meta-Llama-3-70B-Instruct} & $82.0$ & $81.3$ & $8.92$ &$34.4$  & $36.33$ & $3.82(0.128)$  \\ %
\midrule
\texttt{Qwen1.5-72B-Chat} & $75.6$ & $65.5$ & $8.61$ &$36.6$  & $31.33$ & $3.45(0.089)$  \\ %
\midrule
\texttt{Mixtral-8x7B-Instruct-v0.1} & $70.6$ & $57.3$ & $8.30$ &$23.7$  & $44.67$ & $2.02(0.027)$  \\ %
\midrule
\texttt{vicuna-33b-v1.3} & $59.2$ & $52.0$ & $7.12$ &$12.7$  & $21.33$ & $0.35(0.033)$  \\ %
\midrule
\midrule
$\rho$ $\uparrow$ & $0.82$ & $0.71$ & $0.71$ & $1.0$ & $-$ & $\mathbf{0.94}$\\
\midrule
$\tau$ $\uparrow$ & $0.73$ & $0.60$ & $0.60$ & $1.0$ & $-$ & $\mathbf{0.86}$\\
\bottomrule
\end{tabular}
\caption{Results of More Powerful Models. Given the strong capabilities of the evaluated models, we use Yi-34B-Chat as the baseline and exclude the AlpacaEval benchmark, which is relatively simple for these models.} 
\label{tab:more}
\end{table*}

\noindent \textbf{More powerful models.}
To demonstrate the performance of our approach in comparison with more powerful models, Table \ref{tab:more} presents the results of using Yi-34B-Chat as the baseline. We selected some of the most advanced open-source models currently available for testing against Yi-34B-Chat. Our TreeEval achieved performance closest to that of AlpacaEval2. More model results can be found in the (appendix More model results).

\begin{table*}[ht]
\captionsetup{aboveskip=1pt, belowskip=1pt} 
\centering
\small
\begin{tabular}{l | c | c | c | c | c | c}
\toprule
Model & \thead{\texttt{Yi-34B}\\\texttt{-Chat}} & \thead{\texttt{Xwin-LM}\\\texttt{-13B-V0.1}} & \thead{\texttt{Mistral-7B}\\\texttt{-Instruct-v0.2}} & \thead{\texttt{vicuna}\\\texttt{-33b-v1.3}} & \thead{\texttt{WizardLM}\\\texttt{-13B-V1.2}} & \thead{\texttt{zephyr}\\\texttt{-7b-beta}} \\
\midrule
\texttt{Yi-34B-Chat} & $-$ & $1.88(0.400)$ & $1.52(0.010)$ & $2.1(0.070)$ &  $1.21(0.076)$ & $1.75(0.143)$ \\
\midrule
\texttt{Xwin-LM-13B-V0.1} & $3.12(0.400)$ & $-$ & $2.33(0.000)$ & $1.53(0.403)$ & $1.57(0.109)$ & $2.41(0.000)$ \\
\midrule
\texttt{Mistral-7B-Instruct-v0.2} & $3.48(0.010)$ & $2.67(0.000)$ & $-$ & $1.61(0.044)$ & $1.10(0.070)$ & $2.19(0.003)$ \\
\midrule
\texttt{vicuna-33b-v1.3} & $2.9(0.070)$ & $3.47(0.403)$ & $3.39(0.044)$ & $-$ & $2.01(0.374)$ & $3.7(0.071)$ \\
\midrule
\texttt{WizardLM-13B-V1.2} & $3.79(0.076)$ & $3.43(0.109)$ & $3.90(0.070)$ & $2.99(0.374)$ & $-$ & $3.94(0.069)$ \\
\midrule
\texttt{zephyr-7b-beta} & $3.25(0.143)$ & $2.59(0.000)$ & $2.81(0.003)$ & $1.3(0.071)$ & $1.06(0.069)$ & $-$ \\
\bottomrule
\end{tabular}
\caption{Our result for each model pairs. The elements in this table represent the scores obtained by comparing models using treeEval, with the column model being compared against the row model.}
\label{tab:sub_topic}
\end{table*}

\noindent \textbf{Pairwise Comparison for different model pairs.}
We iteratively change the references for the pairwise comparison and the results are shown in Table~\ref{tab:sub_topic}.
Choosing the right baseline model is a critical step in our evaluation strategy. We selected the \texttt{Mistral-7B-Instruct-v0.2} as our baseline, emphasizing the significance of selecting a baseline that accurately reflects the broad insights from pairwise model comparisons. Ideally, a baseline model should have a performance level that is neither too high nor too low, ensuring fair and balanced comparisons across all models. Interestingly, our observations indicate that even a baseline model chosen at random can lead to rankings that closely resemble those from a thorough pairwise evaluation. Thus, it's feasible to start with a randomly chosen baseline model to set up an initial order of performance. This preliminary order can be refined effectively using the bubble sort method. Given the initial order's similarity to the final ranking, this refinement process tends towards an \(O(n)\) complexity, significantly enhancing the evaluation's precision and efficiency.



\noindent \textbf{Ablation Studies.} 
As we can see in Table~\ref{tab:ablation}, changing BFS search to DFS search dramatically increases the number of questions but decreases the performance.
This is because DFS search generates the child node first rather than the sibling node such that the influence of sibling node will be neglected in both question generation and termination identification procedures.
Removing step one, which indicates skip the topic generation step, decreases the performance.
This indicates the significant role of identifying the topic for question generation.
When we iteratively remove the scores in aggregator, we observe general performance drop on $\tau$.
This indicates that all the scores in the aggregator are important in producing a comprehensive score.

\begin{table}[]
\captionsetup{aboveskip=1pt, belowskip=1pt} 
\centering
\small
\begin{tabular}{l | c c c }
\toprule
 Methods & \#Q & $\rho$ & $\tau$  \\
 \midrule
TreeEval & $45.1$& $0.83$&$0.73$ \\
\midrule
BFS $\rightarrow$ DFS & $149.4$ & $0.37$&$0.33$ \\
w/o Step One & $49.3$  & $0.31$ &$0.2$ \\
w/o $w^{\text{root}}$ & $45.1$ &$0.77$&$0.6$  \\
w/o $w^{\text{topic}}$ & $45.1$&$0.77$&$0.6$ \\
w/o $w^{\text{sibling}}$ & $45.1$&$0.71$&$0.47$ \\
\bottomrule
\end{tabular}
\caption{Ablation study on TreeEval.}
\label{tab:ablation}
\end{table}

\noindent \textbf{Case studies} are presented in (Appendix Case Studies).

\section{Conclusions}
\label{sec:con}

In this paper, we introduce TreeEval, a benchmark-free evaluation approach for LLMs with tree planning, which automatically controls the evaluation process with tree planning.
We experimentally verify that TreeEval can not only produce reliable evaluation results without data leakage but also enhance discrimination between similarly performing LLMs.

\section{Limitations}
\label{limit}

Using LLMs like GPT-4 as judges introduces potential data leakage risks due to biases in their pre-training data. This can be mitigated by selecting neutral evaluators independent of the assessed models' training data or randomly rotating evaluators to reduce bias.  

While GPT-4 is a powerful examiner, it has limitations, particularly in areas outside its expertise. This can be addressed by providing more contextual guidance during evaluations. In the future, training specialized evaluators to extract questions from document repositories and assess comprehension could ensure more accurate, domain-specific evaluations.

\section{Ethical Considerations}
\label{ethics}
Although we prioritize the security of the LLMs we use during evaluations, striving to employ aligned LLMs with higher safety standards, and endeavor to ensure that LLM outputs adhere to ethical and legal requirements, limitations arising from model size and probabilistic generation paradigms may lead to various unexpected outputs. These could include questions or responses containing biases, discrimination, or other harmful content. Please refrain from disseminating such content.
\section{Acknowledgements}
\label{sec:ack}

The authors would like to thank the anonymous reviewers for their insightful comments. This work is supported by the National Key Research \& Develop Plan (2023YFF0725100) and Young Scientists Project of National Natural Science Foundation (Project No. 62206097).
\bibliography{aaai25}

\onecolumn
\section{Appendix}
\label{sec:appendix}

\subsection{More analysis}
\label{ap:correlation}

\begin{figure*}[htbp]
\centering
\begin{minipage}{0.45\textwidth}
  \centering
  \includegraphics[width=\linewidth]{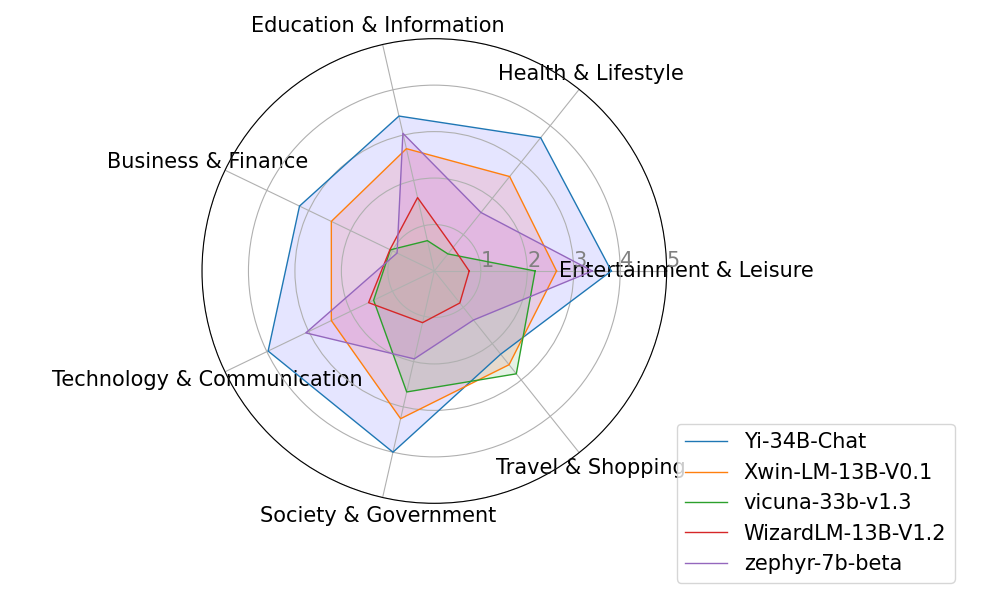}
  \caption{Radar chart illustrating the scores of various LLMs under different pre-defined topics.}
  \label{fig:radar}
\end{minipage}\hfill
\begin{minipage}{0.45\textwidth}
  \centering
  \includegraphics[width=\linewidth]{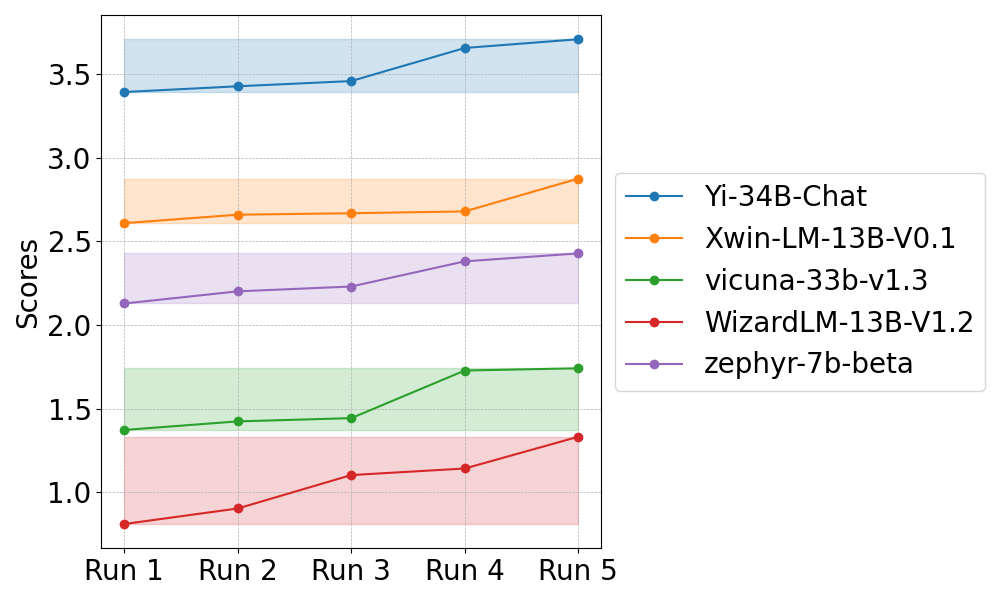}
  \caption{Re-run TreeEval 5 times for various LLMs.
  }
  \label{fig:linechart}
\end{minipage}
\end{figure*}

\begin{figure*}[htbp]
    \centering
    \includegraphics[width=.95\linewidth]{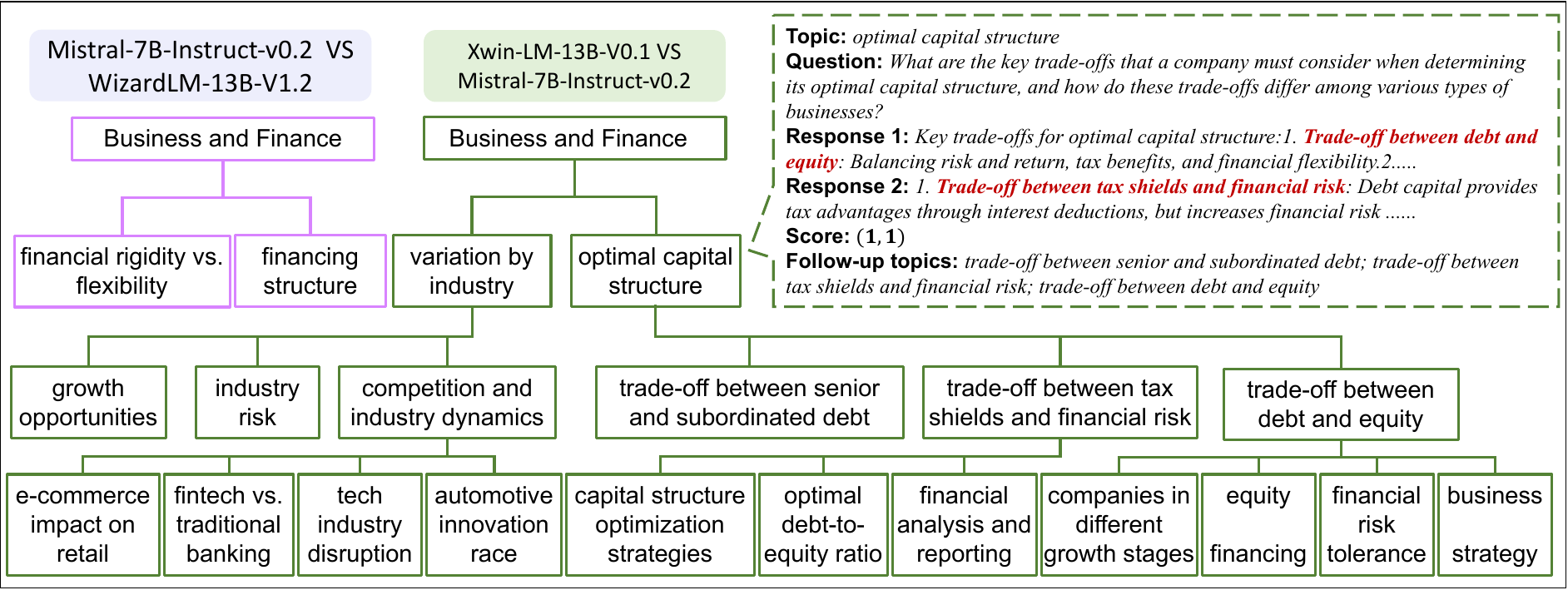}
    \caption{Examples of evaluation process for two pairs of LLMs under topic ``\textit{Business and Finance}'', which are shown in two colored trees.
    The detailed contents of a node is displayed in a dashed box and the recognized entities used for follow-up topics are shown in red fonts.}
    \label{fig:pipeline}
\end{figure*}

\subsubsection{Fine-grained Evaluation.}
\label{ap:fine-grained}
From the Figure~\ref{fig:radar}, we can observe that different LLMs excel in various knowledge domains. 
The performance of the same model may vary across different fields.
For example, \texttt{Yi-34B-Chat} is short in \textit{Travel and Shopping} while it demonstrates relatively good performance on other topics.
Through our TreeEval, we can diagnose an LLM with the fine-grained results in diverse domains.

\subsubsection{Robustness of TreeEval.} 
\label{ap:robustness}
We draw Figure~\ref{fig:linechart} to show the evaluation results of different LLMs in multiple runs. 
We can see that, for a given LLM under evaluation, conducting multiple repeated experiments yields relatively similar scores when the examiner's temperature is set to 1.
The low variance indicates that TreeEval is able to generate stable and robust results.

\subsubsection{Case Studies.} 
\label{ap:case}
In Figure~\ref{fig:pipeline}, it's clear that TreeEval effectively identifies performance gaps between LLMs. For instance, \texttt{Mistral-7B-Instruct-v0.2} notably outperforms \texttt{WizardLM-13B-V1.2}, reflected in a smaller tree. Conversely, when models perform similarly, like \texttt{Mistral-7B-Instruct-v0.2} and \texttt{Xwin-LM-13B-V0.1}, TreeEval constructs a larger tree to discern subtle performance differences.

\begin{table*}[t!]
\captionsetup{aboveskip=2pt, belowskip=2pt} 
\centering
\small
\begin{tabular}{l | c | c | c | c | c | c c }
\toprule
LLMs & \thead{MMLU}$^\star$ & \thead{BBH}$^\star$ & \thead{AlpacaEval}$^\dagger$ & \thead{MT-bench}$^\dagger$ & \thead{AlpacaEval2.0}$^\dagger$ & \multicolumn{2}{c}{TreeEval(Ours)} \\
\midrule
  & Acc & Acc & Win-Rate & score & Win-Rate& \#Q & Score  \\
\midrule
\texttt{Mistral-7B-Instruct-v0.2} & $70.6$ & $46.4$ & $92.78$ & $8.30$ &$14.72$  & $-$ & $2.50$  \\ %
\midrule
\texttt{Vicuna-7b-v1.3} & $47.1$ & $37.5$ & $76.84$ & $6.00$ &$4.64$ & $34.67$ & $0.92$  \\ %
\midrule
\texttt{Vicuna-7b-v1.5} & $49.9$ & $43.3$ & $74.50$ & $6.17$ & $4.79$ & $36.33$ & $0.96$ \\
\midrule
\texttt{Vicuna-13b-v1.3} & $52.1$ & $43.5$ & $82.11$ & $6.39$ &$7.13$ & $38.33$ & $1.28$  \\ %
\midrule
\texttt{Vicuna-13b-v1.5} & $55.8$ & $51.5$ & $81.07$ & $6.57$ & $6.72$ & $34.33$ & $1.06$ \\
\midrule 
\texttt{chatglm2-6b} & $45.5$ & $32.7$ & $47.12$ & $4.96$ & $2.76$ & $39.67$ & $1.12$ \\
\midrule 
\texttt{alpaca-13b} & $55.1$ & $43.3$ & $78.93$ & $6.53$ & $7.43$ & $55.00$ & $2.19$ \\
\midrule 
\texttt{Qwen-14B-chat} & $66.5$ & $58.0$ & $79.83$ & $6.96$ & $7.50$ & $36.67$ & $1.40$ \\
\midrule 
\texttt{Starling-LM-7B-alpha} & $63.9$ & $51.5$ & $91.99$ & $8.09$ & $14.24$ & $52.00$ & $2.36$\\

\midrule 
\midrule
$\rho$ $\uparrow$ & $0.57$ & $0.21$ & $0.25$ & $0.31$ & $1.0$ & $-$ & $\mathbf{0.70}$\\
\midrule
$\tau$ $\uparrow$ & $0.53$ & $0.14$ & $0.20$ & $0.36$ & $1.0$ & $-$ & $\mathbf{0.58}$\\
\bottomrule
\end{tabular}
\caption{More model results} 
\label{tab:more1}
\end{table*}

\subsubsection{More model results.}
\label{ap:results}
To further demonstrate the performance of our method on different models to be tested, we also show the performance of the vicuna series models as well as chatglm2-6b~\cite{zeng2022glm} and alpaca-13b~\cite{claude2-alpaca} in Table \ref{tab:more1}. But the main analysis still focuses on the previous six models.

\subsection{Eval Controller Example}
\label{ap:Controller Example}
For the first aspect: the generated follow-up question should be closely linked to its previous question via
topics. For example, in Figure \ref{fig:method}, we generate $Q_2$ as "How does 5G technology improve Internet of Things (IoT) applications and smart city initiatives?" Based on the $C_6$ we obtained we generated a new question $Q_6$ "In what ways does 5G technology enable advancements in smart home devices and automation?", $Q_6$ is an expansion and extension of $Q_2$ on $C_6$ "applications". 

For the second aspect: the generated questions should not be repeated in the existing questions. For example, in Figure \ref{fig:method}, $Q_5$ was initially "What are the ethical considerations when designing AI systems for human-machine interaction?" However, this significantly overlapped with our previously established $Q_3$ "What are the key ethical considerations when developing AI technologies for communication platforms?" Consequently, we opted for $Q_5$ to be "What role does AI play in the development of voice-activated systems, and how does it change human-machine interaction?" to ensure variety and specificity in our discussion topics. 

\begin{figure*}[t!]
    \centering
    \includegraphics[width=\linewidth]{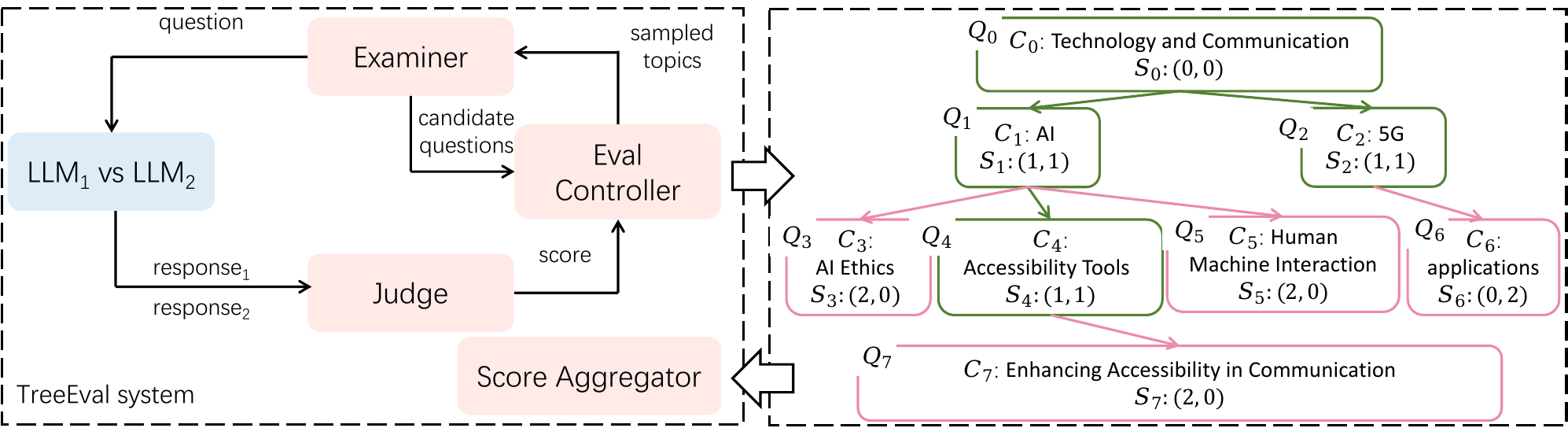}
    \caption{To facilitate analysis, Figure \ref{fig:method} is re-placed here.}
\end{figure*}

\subsection{Prompt for Examiner}
\label{ap:examiner}

I want you to assume the role of the expert and ask a question that expands and reflects your understanding of \{topic\}. Your task is to ask a question about \{topic\}. 
Only through a profound understanding of \{topic\} can one correctly answer this question. 
Please adhere strictly to the following 4 task guidelines:
\begin{enumerate}
    \item Your question should begin with a question word, such as "what", "which", "when", "where", "how", "why", etc.
    \item The objective of your question should be to manifest the respondent's understanding of \{topic\} and to differentiate respondents based on their comprehension level.
    \item Questions should be self-explanatory, not requiring additional context or clarification.
    \item Please format your question in the JSON structure provided below. Remember, only output the content in the following format, and nothing else: \{\{"question": your question\}\}
\end{enumerate}

\subsection{Prompt for Judge}
\label{ap:judge}

You are assessing two submitted responses to a user's query based on specific criteria. Evaluate the quality, relevance, accuracy, clarity, and any other relevant factors to determine which response is superior, or if they are equally valuable or lacking. Here is the data for your assessment:

[Query]: \{question\}

[Response 1]: \{answer 1\}

[Response 2]: \{answer 2\}

Assessment Criteria:
\begin{enumerate}
\item Relevance to the query: Does the response directly address the user's question or concern?
\item Accuracy of information: Are the facts or solutions provided in the response correct and reliable?
\item Clarity and comprehensibility: Is the response easy to understand, well-structured, and free of jargon or ambiguity?
\item Completeness: Does the response cover all aspects of the query or offer a comprehensive solution?
\item Additional value: Does the response provide extra insights, tips, or information that enhances the user's understanding or solves the problem more effectively?
\end{enumerate}
Instructions for Assessment:
\begin{enumerate}
\item Identify and focus on the criteria that significantly distinguish the two responses. Disregard criteria that do not offer a clear distinction.
\item Consider any specific aspects of the query and the responses that may require additional factors for a fair comparison. Mention these factors explicitly.
\item Conclude your assessment by deciding which response is better, or if they are tied. Your decision must be based on a coherent evaluation across the mentioned criteria and any additional factors you've identified.
\end{enumerate}
Please return your final decision in the following JSON format:
\{"Eval\_result": "Response 1"/"Response 2"/"Tie"\}

Note: Remember, the output should only contain the decision in the specified JSON format and nothing else.

\subsection{Prompt for NER}
\label{ap:ner}

You are asking questions and answers based on a topic you know and based on this topic. Please extract some subtopics from the answers. Here's an example: 

Here is the data:

      [Input data]
      
      ***
      
      [topic]: programming languages
      
      ***
      
      [question]: Which programming languages can you write code in?
      
      ***
      
      [answer]: I know python, C++, R language, etc.
      
      ***
      
      [Output Data]
      
      ***
      
      [subtopic] : ["python","C++","R language"]
      
now the official question

Here is the data:

      [Input data]
      
      ***
      
      [topic]:{0}
      
      ***
      
      [question]:{1}
      
      ***
      
      [answer]:{2}
      
      ***
      
      [Output Data]
      
      ***
      
Please return your final decision in list format. Remember, you only need to output the content in the following List format, with each element as a subtopic and nothing else. Remember, you only need to output the three most important subtopics in the following List format.

[subtopic] : ["subtopic1","subtopic2","subtopic3"]

\end{document}